\documentclass{IOS-Book-Article}

\usepackage{mathptmx}
\usepackage{graphicx}
\usepackage{soul}\setuldepth{article}
\usepackage{subcaption}
\usepackage{soul}

\def\hb{\hbox to 11.5 cm{}}

\begin{document}

\pagestyle{headings}
\def\thepage{}

\begin{frontmatter}              

\title{Towards an AI Assistant for Power Grid Operators}

\markboth{}{June 2021\hb}

\author[A]{\fnms{Antoine} \snm{Marot}}, %
\author[A]{\fnms{Alexandre} \snm{Rozier}},
\author[A]{\fnms{Matthieu} \snm{Dussartre}},
\author[A]{\fnms{Laure} \snm{Crochepierre}}
and \author[A]{\fnms{Benjamin} \snm{Donnot}}

\address[A]{RTE, R\&D and AI Lab, France}

\begin{abstract}
Power grids are becoming more complex to operate in the digital age given the current energy transition to cope with climate change. As a result, real-time decision-making is getting more challenging as the human operator has to deal with more information, more uncertainty, more applications, and more coordination. While supervision has been primarily used to help them make decisions over the last decades, it cannot reasonably scale up anymore. There is a great need for rethinking the human-machine interface under more unified and interactive frameworks. Taking advantage of the latest developments in Human-Machine Interface and Artificial Intelligence, we expose our vision of a new assistant framework relying on an hypervision interface and greater bidirectional interaction. We review the known principles of decision-making driving our assistant design alongside with its supporting assistance functions. We finally share some guidelines to make progress towards the development of such an assistant.
\end{abstract}

\end{frontmatter}
\markboth{June 2022\hb}{June 2022\hb}

\section{Introduction}

From the beginning, power grids have been complex artificial systems. As of today, complexity keeps rising at a time of energy transition given the advent of intermittent renewable energies on the production side and of prosumers on the demand side, coupled with the globalization of energy markets over a more and more interconnected European grid. Operators are facing aging grids, with slower grid asset developments due to decreasing social acceptability. Operators hence need to operate a system closer to its limits while dealing with greater uncertainty and increasing grid automation \cite{marot2022perspectives} inducing complex cyber-physical dynamics \cite{allgoewer2019position}. 
While there could eventually be a temptation to develop a fully autonomous grid to cope with that complexity, it falls short for such large critical system operations. Indeed, coordination, responsibility, accountability, and explainability are a must when operating such a system and can only be reasonably achieved by humans today: human operators remain key players \cite{prostejovsky2019future}.


Historically, control rooms have tackled grid evolution dynamics by gradually adding more applications and screens in control rooms. While incrementally growing a monitoring alarm-driven management system has been relatively effective until now, we are today struggling to maintain effective decision-making at a new scale of complexity, essentially in terms of manageable cognitive load for human operators, who should remain at the center of decisions \cite{marot2022perspectives}. In that regard, Human-Machine Interfaces (HMI) have been identified as a risk factor for human error \cite{flaspoler2009human}, and should now be considered more closely. In addition, several decades of research in psychology \cite{kahneman2011thinking} and neuroscience \cite{naccache2005effortless} have shed light on the human decision-making process and its limits, which could in turn lead to improving it.


Lately, several works have proposed situation awareness frameworks \cite{naderpour2014intelligent,panteli2015situation,liu2016situational} to help augment the operator's comprehension of safety-critical situations. In addition to better information processing, it is also urgent to rethink the operator's human-machine interface \cite{naderpour2014intelligent} and interaction to assist the operator's regular real-time decision-making. Rather than having operators adapt to the machine through a technology-centered system engineering design, machines and operators could co-adapt \cite{mackay2000responding} following a more human centered-design \cite{boy2017human,shneiderman2022human} approach, possibly rooted in the older concept of man-computer symbiosis \cite{licklider1960man}.
In terms of interfaces,  we have seen tremendous innovations in other domains, especially in consumer products such as smartphones, connected homes, social networks, search engines, and recommendation systems. They have been well-adopted for ergonomically providing the most relevant information to the user, mostly on single screens, while dealing in the background with vast amounts of information. In the future, the webstrates \cite{klokmose2015webstrates} concept might enable very adaptive, evolving, personal and consistent interfaces by sharing media ubiquitously through devices and applications.

The development of today's interfaces has also been made possible thanks to the latest developments of Artificial Intelligence (AI), and particularly in the field of Machine Learning (ML). These advances enabled deeper and more practical large-scale real-time information processing, such as in computer vision \cite{russakovsky2015imagenet}, image understanding \cite{hossain2019comprehensive}, natural language processing \cite{brown2020language}, and recommendations \cite{zhang2019deep}.
This shows a shift towards even more advanced HMI, 
through the concept of assistants. Assistants were found useful to both improve single-user performance and group collaboration on a common task \cite{winkler2019alexa}, help chess players develop their playing skills \cite{mcilroyyoung2020maia}, or help programmers write code \cite{chen2021evaluating}. In power grids, the notion of an AI assistant was used lately  in \cite{donnot2017introducing,marot2018expert} and listed as an opportunity to tackle climate change \cite{clutton2021climate}. Nevertheless, eventual assistant pitfalls should be closely investigated, analogous to the very influential ``ironies of automation``, which notably used power grids as an introductory example \cite{bainbridge1983ironies}. 

In this paper, we present the future design of operator AI-infused assistant given the latest developments in HMI, AI, and decision-making science. Our contributions lie in:
\begin{itemize}
    \item establishing links between these different domains and our application scenarios to push towards appropriate solution design while being aware of pitfalls;
    \item broadening the perspective to other research fields related to human behavior and cognition beyond the usual situation awareness focus in power systems \cite{prostejovsky2019future, panteli2015situation};
    \item adapting existing AI assistant framework and guidelines to power grid specificity; 
    \item proposing hypervision interface as an effective hybrid decision-making enabler.
\end{itemize}
We first describe in Section \ref{sec:use_case_description} the industrial use case we focus on. Then in Section \ref{sec:humanai_dynamics_in_decision_making}, we review the effectiveness and limits of humans and AI in decision-making, highlighting their complementarity. We further define an assistant in Section \ref{'defining-an-assistant'}. We then discuss in Section \ref{sec:expected_functions} the specificity of an assistant for grid operators. Finally, in Section \ref{sec:guidelines}, we devise guidelines for how researchers can develop and test such an assistant. 

\section{Use case description}
\label{sec:use_case_description}
We are here considering the field of evolving power grid operations for the Energy Transition. Operations have to adapt as the ongoing decarbonization to cope with Climate Change modifies the generation, electrifies the demand, and digitizes system devices. 

\subsection{Today's operations}

Today grid operators operate in real-time from control rooms to optimize the power flows on electrical lines, handle maintenance with planned outages or new equipment integration on the grid, and most importantly, avoid blackouts because of congestion. More details about their role and tasks can be found in \cite{prostejovsky2019future}. They are highly trained engineers as their job requires studies, planning, and adaptable decision-making rather than simply reproducing pre-established  event management scenarios. They operate based on simulation tools, real-time and forecasted data, but yet with little decision-making support tools such as assistants.  When they feel they need to solve a problem, they mostly manually explore solutions and validate their decision in their simulation tool. They can modify the line connectivity on the grid to reroute power flows, but also modify some production, limit consumption by a few percent, or even use battery storage today to change the power flows on the grid. This is a large set of possible flexibilities among which they have to identify the effective ones in a given context. Day-ahead planning services give insights on the upcoming trends and possible actions to start considering if some problem occurs, in addition to outage planning. There then exists an intra-day established workflow with 5-minute times step forecast resolution over a few hours' horizons.  But this requires a lot of supervision, a lot of manual entries and manual simulations. They operate mainly with experience to determine relevant remedial actions. 

As the variability on the grid is increasing a lot with shifting dynamics and behaviors even within a year now given the energy transition, usual solutions might not work anymore in all or new contexts. Operators will have to adapt more quickly, leveraging human flexibility if given a proper work environment \cite{abram2021flexibility}. As it can take months today for an operator to gain the necessary knowledge on a different system through extensive training through manual studies, new assistance will be needed, as much as improved performance evaluation \cite{teive2022intelligent}. Also, while some flexibilities and actions can be leveraged curatively once a problem occurs, sometimes actions have to be taken preventively before it is too late to implement them given operational constraints, in particular when considering using batteries or redispatching.
Several operators also act on the same interconnected grid at the same time, and this requires some coordination that is not always easily achieved despite common grid representations.

\subsection{Current limitations today and assistant need}
The current ``supervise everything'' approach on dozens of screens cannot scale anymore to take critical decisions as the complexity rises: operators become overwhelmed by information without many indications on what to do and any recommendations for it. The volume of data to consider only gets increasing by 2 to 3 orders of magnitude to keep the ability to predict and anticipate in a more stochastic and constrained system. There is hence a need to help operators identify and prioritize tasks while displaying relevant information and recommendations only for those tasks, with ideally a single interface assistant. In particular, usual approaches from operational research \cite{bhaskar2011security,heidarifar2021optimal} don’t leverage very well available historical data, whereas current data-driven approaches could provide a welcome speed-up. They don’t integrate operator experience and preferred thought process, hence limiting their acceptability. New research is however emerging to better integrate human decision-making \cite{mohrlen2022humans}.
Finally, current workflow only considers snapshot-based solutions and not sequential decisions over a time horizon, thus putting the burden of the anticipation and evaluation of the long-term consequences of the actions on the human operator alone. 

As the grid gets pushed towards its limits, decisions become more numerous and a lot more interdependent. Solutions should not only be effective at one given time but over a larger time horizon. Solutions should also be implemented with the right anticipation given their effective activation time: switching line connectivity is quick but starting production can sometimes take a few hours. So decisions will need to indeed consider the full underlying planning problem of power grid operations (as illustrated in section \ref{sec:use_case_example}). Better near real-time operation planning assistance could also help anticipate the workload and level it up across operators in a given control room.

An assistant could hence help augment the operator's decision-making capability  to address \cite{marot2022perspectives}: (a) More rapidly evolving and changing system environment; (b) More numerous, complex, and coordinated decisions to make; (c) More uncertainty to consider and more anticipation needed; (d) Overcrowded and fragmented work environment with multi-screen applications and data patchwork; (e) Human operator cognitive load saturation. It would eventually help limit the grid operational cost to a few percent increase rather than a 2-fold one. The ambition is to use at least twice as much flexibility as of today, i.e.\ to use them more frequently and with more diversity \cite{energy2021pathways}. It would also facilitate workload \cite{ghalenoei2022impact} management across teams for improved coordination \cite{harbers2017value} while avoiding cognitive load saturation, and deal with information overload. The interactions should not be only limited to isolated Operator-Assistant pairs, but involve teams of interrelated activities through a shared information and task management system. It should however not come at the cost of decreasing responsibility and accountability, making clear who is in charge and providing necessary interpretability for operators to make explanations.


\section{Human-AI dynamics in decision-making}
\label{sec:humanai_dynamics_in_decision_making}
\subsection{Human decision-making} 
Human decision-making  is a matter of  attention and executive control \cite{naccache2005effortless}. Taking proper decisions first involves paying attention to the right information in the environment and making sense of it. It further implies selecting relevant actions while inhibiting inappropriate ones and eventually executing one in a timely manner. 
Following dual-process theory in the psychology of human reasoning \cite{kahneman2011thinking}, already successfully applied to domains such as medical decision-making \cite{djulbegovic2012dual}, we can describe two underlying imaginary operating and cooperative agents, called System 1 (S$_{1}$) and System 2 (S$_{2}$). S$_{1}$ is the fast intuitive and heuristic agent while S$_{2}$ is the slow and reasoning agent. S$_{2}$ is the one responsible for decisions assisted by S$_{1}$ which continuously and automatically provides him predictions for action. Most of the time, S$_{2}$ just lazily accepts S$_{1}$ proposal in usual situations without much more thinking, resulting in successful, quick and cognitively effortless decisions. When confronted with unusual situations, however, S$_{2}$ can develop more explicit conscious thinking, beyond S$_{1}$ predictions, to deliberate and come up with novel and acceptable decisions while cognitively costly. 

Young operators will rely more heavily on S$_{2}$ and can hence struggle to take any good decision on time for several situations that still appear complex and unusual to them. As they are very focused on trying to make sense of it, they have narrow attention and might miss important new information. As they learn overtime through appropriate training and feedback become expert, climbing through the expertise ladder \cite{stevens2016defining}, their intuitive S$_{1}$ grows for that field of expertise. This enables them to make good and quick decisions even more often with ease. For an expert operator, it has become a lot easier to operate a system intuitively, being able to make more decisions as well as decisions in more difficult situations \cite{EPRI2021}. However, the downside can be overconfidence, overlooking unusual information that would require more deliberation from S$_{2}$.

\subsection{Human biases and desirable assistance}
\label{sec:human_biaises}

Because it relies on fast heuristics and mostly jumps to conclusions, System 1 indeed introduces several potential biases which can lead to human errors, hence limiting the effectiveness of human decision-making. Cognitive biases are summarized in the cognitive bias codex \cite{codex}
and classified through 4 problems they are trying to circumvent: a limited memory, the need to act fast, the information overload and a lack of meaning. Among possibly damaging biases, we can more specifically list:
\begin{itemize}
\item anchoring bias: be over-reliant on the first piece of information we see.
\item confirmation bias: paying more attention to information confirming our beliefs.
\item overconfidence bias: too confident about one's abilities leads to greater risk-taking.
\item information bias: tendency to seek information when it does not affect action (more information is not always better). 
\item availability heuristic: overestimate the importance of information that is available.  
\item ostrich effect: ignore dangerous or negative information. 
\item outcome bias: judge a decision based on the outcome rather than how it was made.
\end{itemize}

An assistant should hence help the operator avoid biases through assistance in:
\begin{itemize}
\item augmenting his memory, knowledge retrieval and keeping track of latest events.
\item better information filtering or highlighting, enhancing attention focus.
\item contextualizing a situation and giving feedback.
\item making recommendations, possibly handling some tasks or alerting on undesirable expected consequences.
\end{itemize}

Let's now consider what AI in its latest developments could bring in that regard.

\subsection{AI potential for assistance}

The recent deep learning revolution demonstrated some impressive practical abilities of AI by being able to digest a lot of information, memorize large historical datasets, and learn by imitation to infer quickly effective actions in context. Turing-award Yoshua Bengio recently described current deep-learning AI as a S$_{1}$ kind of intelligence \cite{bengio2019system}, while missing S$_{2}$ reasoning. This type of AI is  presented as advanced pattern matching and recognition machines like S$_{1}$ \cite{kahneman2011thinking}, being coined as artificial intuition \cite{perez2018artificial}. It however lacks the ability to reason about causality \cite{pearl2018theoretical}, hence lacking understanding and common sense. Yet Human and AI can be seen as complementary heterogeneous intelligence that could achieve a superior outcome when co-adapting \cite{mackay2000responding} and developing hybrid intelligence \cite{dellermann2019hybrid} or human-centered AI \cite{shneiderman2022human}. This is best exemplified by Kasparov's ``Centaur chess`` concept \cite{case2018become}, where humans play alongside machines to reach superior performances. When powering assistance systems, AI seems capable of overcoming some previously mentioned human limitations by enhancing S$_{1}$ operator's ability, whose S$_{2}$ remains in charge of final decisions. Some initial assistance for S$_{2}$ can nonetheless be considered through explanations or counterfactual reasoning if a simulator exists.   

\subsection{Pitfalls of Human-Machine hybridization}\label{sec:pitfalls}

As we have seen in the previous section, human-AI partnerships shows many promises. But such systems must be carefully designed, as numerous shortcomings might arise. We can make the analogy to "ironies or myth of automation" \cite{bainbridge1983ironies,bradshaw2013seven} which warns that operators should ultimately be more skilled than less skilled, and less loaded than more loaded to deal with the most difficult new and complex situations. Amongst many shortcomings mentioned in  \cite{machines-as-teammates}, over-reliance and deskilling are one of the most damaging for control rooms. As of today, operators have a deep knowledge of their grid area, both in terms of infrastructure and electrical phenomena. Adding an automatic system with persuasive recommendations could lead the operating staff to rely too heavily on the system, progressively losing their expertise, with potentially catastrophic consequences when it fails. Overconfidence should not be a trade-off with over-reliance, and operators should have the opportunity to keep developing their skills and cognitive strategies through regular training and manual problem-solving \cite{bainbridge1983ironies}. 

Additionally, an assistant could misunderstand user intent, and hinder his actions in an example of perverse instantiation \cite{goertzel2015superintelligence}.
These limitations arise from the fact that, while human Systems 1 and 2 are fully integrated into a single cognitive ensemble, AI and humans are separated entities, where communication is key. System 1 works on learned representations of reality, not directly on reality itself, meaning that situations can be interpreted differently. Hence, reference representations should at least be shared in some form between agents to properly communicate, as with language. Not an easy task knowing that, for instance, in language, common words can sometimes have different meanings between people, highlighting not always properly shared representations \cite{marti2019same}. 

Finally, we should resist the temptation of developing an anthropomorphic assistant \cite{mumford2010technics}, i.e.\ not mechanically mimicking humans and human-human interactions. We aim at augmenting the operator, not replacing him, by recognizing the complementary differences between humans and computers \cite{licklider1960man}, and leveraging them for humans.
These motivate the need for design guidelines, which we will define in sections \ref{sec:expected_functions} and \ref{sec:guidelines}, including carefully-crafted human-machine interactions \cite{blanchet:hal-03113458}, or interpretable \cite{doshi2017towards}, explainable \cite{adadi2018peeking} and trustworthy \cite{european2020white} AI. We aim at offering operators high control level despite increasing grid automation without compromising reliability and safety \cite{shneiderman2020human}.

Moving away from these maturing fields that should prove useful to create an assistant, let's define this concept more precisely for our power grid application domain. 


\section{Defining an artificial assistant for power grid operators} \label{'defining-an-assistant'}
We formalize an assistant as an artificial agent providing help on a subset of user tasks within its daily activity, with the final goal of increasing task efficiency while still developing operator skills.


\subsection{Assistant: balancing assistance, user control and automation}
To make things clearer, we should explain how the concept of assistant articulates with assistance functions and is distinct from a completely autonomous system. Assistance functions help users with domain-specific tasks, for example, by alerting when relevant information arrives, or monitoring user context and warning him about unforeseen risks. Situation awareness \cite{panteli2015situation, prostejovsky2019future, naderpour2014intelligent} offers, in that sense, advanced assistance functions. 

Other similar industrial sectors evoke those functions through different autonomy levels \cite{barabas2017current, Brandenburger2019On, frohm2008levels}, that we can reflect on.
We here focus on the Grades of Automation (GoA) definition from the International Association of Public Transport \cite{cohen2015impacts, Brandenburger2019On}:
\begin{itemize}
	\item GoA0: Manual operation with no automatic protection
	\item GoA1: Manual operation with automatic protection
	\item GoA2: Semi-automatic operation
	\item GoA3: Driverless operation
	\item GoA4: Unattended operation
\end{itemize}

GoA1 and GoA2 offer assistance functions as discussed previously, but without much consideration for HMI. GoA3 and GoA4 are targeting autonomy through automation. It highlights that many fields such as autonomous driving \cite{barabas2017current} aim at fully autonomous systems. They diverge from our goal of augmenting operators through an assistant. While being conceptually closer to GoA2, we move away from the usual automation typical of system engineering, to be rather operator centric. 

\begin{figure}[htbp]
\begin{subfigure}{0.8\textwidth}
\centering
\includegraphics[width=\textwidth]{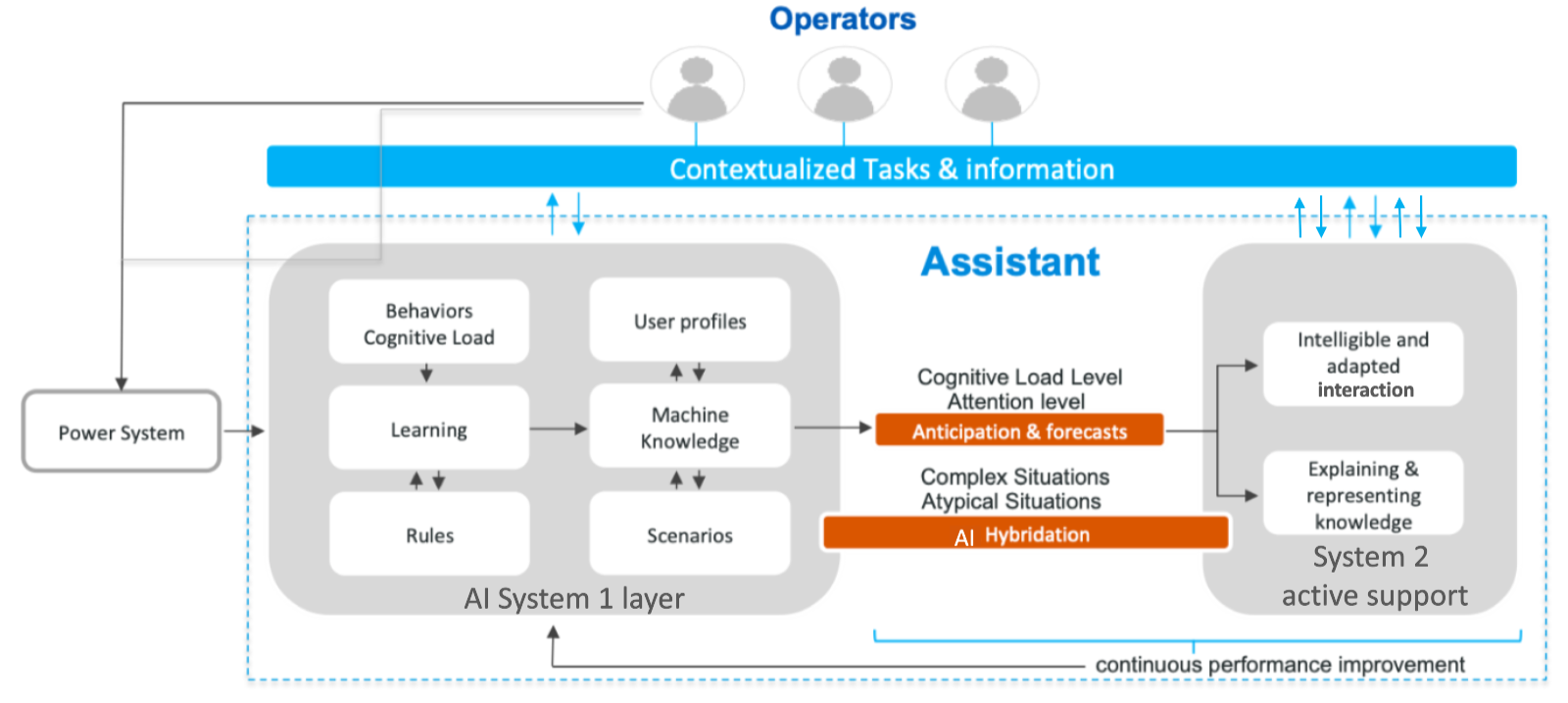}
\caption{The grid operator AI assistant} 
\label{fig:assistant}
\end{subfigure}
\begin{subfigure}{0.18\textwidth}
\vspace{0.8cm}
\includegraphics[width=\textwidth]{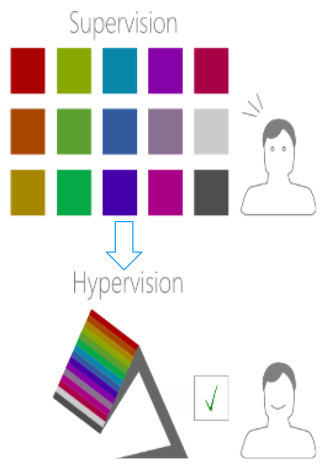}
\vspace{0.1cm}
\caption{Supervision to Hypervision}
\label{fig:Hypervision}
\end{subfigure}
\caption{(a) The grid operator AI assistant relies on Hypervision interface represented as a blue bus, as well as underlying bidirectional interaction and AI components (see section \ref{sec:AIModules}) running altogether in a coordinated and modular fashion. Multiple Operator/Assistant pairs can also coordinate and collaborate through the Hypervision bus. Zooming in, Assistant System 1 type block helps in usual tasks or situations, exploiting core knowledge and intuition learnt. Right block is more dynamic and interactive, hence more bidirectional, to assist in situations that require more reasoning, focus, deliberation or exploration by operators in a more System 2 type fashion. Continuous revision is important to update shared representations. (b) Supervision to Hypervision in \ref{fig:Hypervision} allows moving away from alarm monitoring over many applications on screens (colored squares), to refocus the operator on task completion on a unified but contextualized single interface (mixed colors of merged information from background apps). 
}

\label{fig:full_assistant}
\end{figure}

An assistant, as we illustrate in Fig.\ref{fig:assistant} 
and later discuss, is yet another level whose goal is to offer the right balance between user control and autonomy \cite{heer2019agency} for enhanced decision-making. It certainly has assistance functions at its core, but most importantly also engages actively with the user. It offers a unified interface and allows for dynamic bidirectional interaction with the user to cooperate efficiently on task completion and with others. It is essential to keep in mind that, in this paradigm, the responsibility of system management still ultimately falls on the operators of critical systems. While some tasks could eventually be automated away, it is paramount to avoid pitfalls such as operator deskilling (see section \ref{sec:pitfalls}), often due to the ``out-of-the-loop" effect \cite{prostejovsky2019future}. Assistants should rather let operators remain in control, which will require more than automation supervision \cite{baileytraining}, and help reinforce their expertise, as emphasized in section \ref{bidirectional-interaction}. Finally, unlike GoA framework, getting a teamwork perspective is important, by taking into consideration the interactions between control room operators, to foster proper coordination and to allow for observability, predictability and directability \cite{johnson2018tomorrow}. 

We now focus on core building blocks in creating such assistant, namely hypervision and bidirectional interaction. Applied AI bricks examples would be found in section \ref{sec:AIModules}.

\subsection{Hypervision: smart interface \& information management}\label{hypervision}

Today's supervision leaves to the user the cognitive load to prioritize, organize, and link every displayed information and alarms consistently before considering any decision. It can be regarded as a fragmented ecosystem from an operator's viewpoint. While it has been manageable for up to ten applications, it becomes impractical with always more information and uncoordinated applications to control under heterogeneous formats. Supervision gives access to the user to every information available without much more processing. However, it does not help deal with the information overload and ``lack of meaning'' problems (see \ref{sec:human_biaises})  that need to be tackled for improved decision-making: it dilutes the operator's attention. Let's recall that humans can mainly focus on one task at a time, with a limited working memory space of few information chunks to manipulate \cite{cowan2001magical}. 

To be effective at continuous decision-making, it is often important to focus on the highest priority task at a time, and present only the most relevant information to it. Yet, pausing on one task, dealing with another, and eventually completing the previous one should remain be possible. In that regard, we propose to rely on an ``hypervision" framework to bring the right information at the right time to the right person \cite{marot2022perspectives} while keeping track of user progress for each task. It helps overcome multiple biases, such as both information bias and anchoring bias, by taking advantage of them rather than being influenced by them. 
Hypervision as presented in \cite{marot2022perspectives} relies on the definition of tasks created by processing and synthesizing the necessary information. Those tasks do not have to be only-real time. Indeed, they are still preferably the ones anticipated to be completed or configured ahead of time thanks to forecast, hence defining an expected trajectory that might be adapted along the way. Reaching this higher level of information enables the assistant to establish a simplified but relevant dialogue with the operator, eventually providing him with diagnostics or even recommendations on solutions. Hence, hypervision's goal is to help refocus the operator on task completion rather than alarm monitoring, as illustrated in figure \ref{fig:Hypervision}. While the assistant could instantiate tasks for anticipated risks by the system and help in prioritizing them, the operator should remain free to create and modify some himself, or manually change the order of priority. This manual editing is also important to correct overtime any selection bias that could have been introduced by automatic prioritization, among others. Finally, such tasks can be easily shared and completed between operators, allowing for easier team workload management \cite{harbers2017value}. These create the basis for more advanced and effective bidirectional interaction under shared representations of tasks \cite{heer2019agency} thanks to which users and assistants could work in tandem or in teams to achieve a common objective.


\subsection{Bidirectional interaction}\label{bidirectional-interaction}
While the choice of the interaction modalities (visual, audio, haptic, etc.) matters for enhanced human-assistant partnership, we will leave it open here and focus on the importance of bidirectional interaction between an operator and its assistant.

Put in the spotlight in the 80’s, expert systems developed with AI under predefined rules raised concerns about its practicality for human users in terms of Human-Machine interaction.  Lucy Suchman \cite{suchman1987plans} shed light on their ineffectiveness, mostly attributed to the lack of well-designed interaction and learning loops beyond knowledge retrieval. She noted that plans, similarly to predefined rules, are not  prescriptive  and  not  something  to  follow  exactly, because everything eventually depends on circumstances and contingencies. Plans should rather be seen as heuristic and available resources for actions that help focus one's attention while abstracting the details. But they should get updated through interaction to take proper decisions. In the end, interfaces should not draw a dry delimitation with their user but re-configure themselves and conform with him.

Research \cite{horvitz1999principles}  has shown an increased efficiency in Human-AI coupling when both agents were able to initiate and respond to interaction. These were historically mostly unidirectional, the assistant either asking a predefined set of questions to build its context representation or the user asking to perform some predefined tasks. Such badly designed assistant such as Clippy \cite{baym2019intelligent} could in the end disrupt the user, making it inefficient and frustrating.
In a bidirectional relationship, the interaction is collaborative, with neither the system nor the user in control of the whole interaction \cite{heer2019agency}. The assistant is capable of interacting with the latter to refine its context representation (e.g. ask for a clarification when ambiguities arise), thus improving its efficiency when asked to perform a specific task. A good example of such bidirectional interaction is found in \cite{blanchet:hal-03113458}, where interaction between load-carrying robots and their human partners is learnt over time, resulting in better task completion through operator improvement and per-user adaptation of the robot's behavior.
Further approaches let an assistant learn how and when to defer to an expert \cite{mozannar2020consistent} when in doubt, or conversely let a user interactively learn about the weaknesses and strengths of the assistant under sources of uncertainties \cite{sanchez2022deep}. An assistant can also gently challenge or guide a user to reach novel solutions and build knowledge altogether \cite{liu2017bignav,malloch2017fieldward}. Ultimately a user would like to grasp the underlying assistant model \cite{conversy2018vizir}, to know what the model captured, what he can do with it and how to direct or correct it. Interpretability is at play more than post-hoc explanations \cite{rudin2018please}. In that regard, prospective design \cite{shneiderman2022human,hull2003everything} through interaction and exploration should prove effective, removing the need for the assistant to always explain himself to the operator. An after-operations review process could prove useful for strengthening the relationship \cite{dodge2021after}.

These advances show that creating true human-computer partnerships \cite{beaudouin2021generative} based on the concepts of Discoverability, Appropriability and Expressivity become a reality as well-demonstrated in \cite{koch2020imagesense}. This calls for more academic and industrial collaboration like the ``Cockpit and Bidirectional Assistant" project \cite{cab_project} for critical systems.
In the next section, we review some design guidelines to initiate such partnerships in our use case. 

\vspace{-0.5cm}
\section{Featured function for effective assistant support}
\label{sec:expected_functions}
After defining \emph{what} an assistant should look like in the case of power systems,  we now propose to specify \emph{how} it should proceed when interacting with a user. To do so, we build upon the human-ai guidelines defined by Amershi et Al. \cite{human-ai-interactions} to feature import functions for such assistant. As these guidelines were initially defined from studying consumer products, we highlight some additional specificity when considering industrial systems. 

\textbf{Show contextually relevant information at the right time} - Grid operators evolve in a time-constrained environment where having the right information at the right time is paramount. For instance, power lines reconnected after a routine outage operation for maintenance should be notified as soon as possible to increase grid robustness. An assistant should engage in interaction when the context allows it, taking into account when possible the operator's mental availability and current task, and the expected impact of the interaction. Task prioritization is also a cornerstone of good grid management, and users should be provided with high granularity, curated task details for fast criticality assessment. All of the above-mentioned concerns call for efficient knowledge management and representation, as instantiated by our hypervision system defined in \ref{hypervision}.



\textbf{Scope services and inform the user when in doubt} - 
Doubt can happen when the assistant is uncertain about the user's goal, but also in our case because of assistant model limitations (imperfect grid simulator) or uncertainties in the system (poor weather forecast). Operators are often dealing with variability, be it when anticipating potential hazards due to volatile renewable energies or exploring the effects of preventive actions. 

Let's zoom on the ``Anticipation \& forecasts" slot of fig. \ref{fig:assistant}. When facing uncertainty, for example  when provided with a highly volatile wind forecast, the operator can first simply indicate his intention of monitoring more stable, aggregated regional forecasts and let the assistant switch to a higher resolution as operations shift closer to real-time and more precise weather information arrives. This example also shows that uncertainties when providing assistance should be jointly presented with their probable causes (missing data, poor forecast...), to help decision-making and reinforce operator trust.




\textbf{Support efficient invocation and dismissal} - The number of actions an operator can do in a time window is limited. Interacting with an assistant has to be straightforward. Verbosity should be kept to a minimum, and function of the operator context. In tense situations (e.g.  volatile renewable production), the assistant should initiate interactions more frugally, be more succinct, and support faster dismissal. Knowing how to adapt interactions is not straightforward, as it involves capturing a lot of implicit contextual hints, and getting operator feedback through dialog should play a major role.

\textbf{Take into account and learn from user behavior and feedback} - Grid operators are well-trained experts, capable of evaluating the assistant's answers and providing feedback. Thus, to ensure continuous performance improvement (Fig. \ref{fig:assistant}), the latter must be able to learn from users, for instance by understanding that additional context needs to be considered alongside a specific action, or remembering that a line is under maintenance during a user-provided period. Remembering recent interactions is paramount to user acceptance (nobody likes repeating requests) and capturing user context. 

Moreover, each operator has a personal decision-making style, some relying on numerous power-flow simulations to assess a situation, others relying on their expertise. A good assistant should adjust to these user-specific profiles. 

\textbf{Convey the consequences of user actions} - Not only should an assistant avoid operator deskilling for what was already done effectively, but also should it reinforce user expertise through interaction. In power systems, it is often deplored that the consequences of grid operations are poorly monitored, which in return deprives operators of valuable feedback. Assistants delivering a detailed report of how the grid evolved after a specific action would tremendously speed up the way operators acquire experience by mobilizing their deliberative thinking process (S2 in section \ref{sec:humanai_dynamics_in_decision_making}), and yield better grid management. As what KPIs should be tracked is task-dependent, the operator is also implied in this bidirectional dialog. Logically, interpretability and explainability are inseparable from building this successful human-AI hybridization (fig. \ref{fig:assistant}).


We have seen that human-assistant partnership could be built upon hypervision (\ref{hypervision}), bidirectional interactions and a set of key features. We will now propose some guidelines on more concretely developing, implementing and testing such an assistant.
\vspace{-0.5cm}
\section{Guidelines for developing, implementing and testing an assistant}
\label{sec:guidelines}
Designing an assistant in practice might still seem complex beyond the discussed framework and principles. We devise here some pragmatic guidelines to start simple on a common but modular ground, listing some already available building blocks as well.

\subsection{Grounded Design Considerations}
\vspace{-0.2cm}
\subsubsection{Tasks and Visualizations as shared representations}
In smart grids, functions have been presented in \cite{Li2010_vision, gopstein2021nist,marot2022perspectives} and  tasks described at a high level in \cite{prostejovsky2019future,abram2021flexibility} or through a detailed example \cite{EPRI2021}. In other industrial sectors such as aeronautics \cite{boy2017human}, tasks in processes have been codified more precisely at a granular level
with ontologies \cite{reiss2006using} or conceptual designs from human-computer design \cite{boy2017human}, which gives the operator a clearer framework to work and coordinate with, as for the assistant. 

A task is first defined by the problem to solve specifically, such as a safety problem - a line overload, its priority and the residual time to complete it. It should contain relevant context to understand the root of the problem, what might be already known about it, recent related events or tasks, as well as the persons involved. Building on causal and counterfactual models \cite{pearl2018book} is desirable. A task should further come with suggestions about available actions, and their expected effectiveness. It should finally retain a decision for completion and meta-attributes about it. Structuring tasks this way would allow shared representations \cite{heer2019agency} between the operator and the assistant. Task categories and attributes should be more exhaustively drawn through future works. Eventually, comprehensive activity studies involving multiple tasks in time \cite{boy2017human} should be run.

Also, unlike traditional approaches in power systems that mainly tries to focus on the most critical situations we ever have to solve, we suggest here to start studying tasks in regular situations and gradually increasing the number of needed bidirectional interaction. To operators, it should prove useful to start experimenting on the most routine but sometimes time-consuming tasks with often low added value \cite{endsley2018level}. That way, building trust in the first place should be easier while still helping ease their cognitive load.

Additionally, powerful and interactive visualizations are a common and much appreciated approach for operators to support shared representations and task completion. They are useful companions of an assistant. A recent survey in power grids advocates for such new developments 
\cite{fischer2021towards}. Effective superimposed forecasted and current system state visualization \cite{prouzeau2016towards}, dynamic and temporal network exploration \cite{bach2013graphdiaries,bach2015networkcube}, and high-dimensional event-based visualizations \cite{fouse2011chronoviz} could prove beneficial in that regard.

\subsection{Simple situational use case as a sandbox}
\label{sec:use_case_example}
We offer a simple interesting use case to highlight key difficulties in daily grid management through the interplay of preventive and corrective decisions under uncertainty. This makes us think about how the operator-agent interaction should take place.
An operator starts monitoring a two-area grid at 7:00am. Forecasts show 2 issues:
\begin{itemize}
    \item An incident in area 1 could happen around 9:00am and would lead to some overloads, with three available corrective actions after simulations.
    \item Another such incident in area 2 could happen around 8:30am with only one preventive action available. This leaves only a couple of minutes to execute it or not.
\end{itemize}

\begin{figure}[htbp]
\centering
\includegraphics[width=0.8\textwidth,height=4cm]{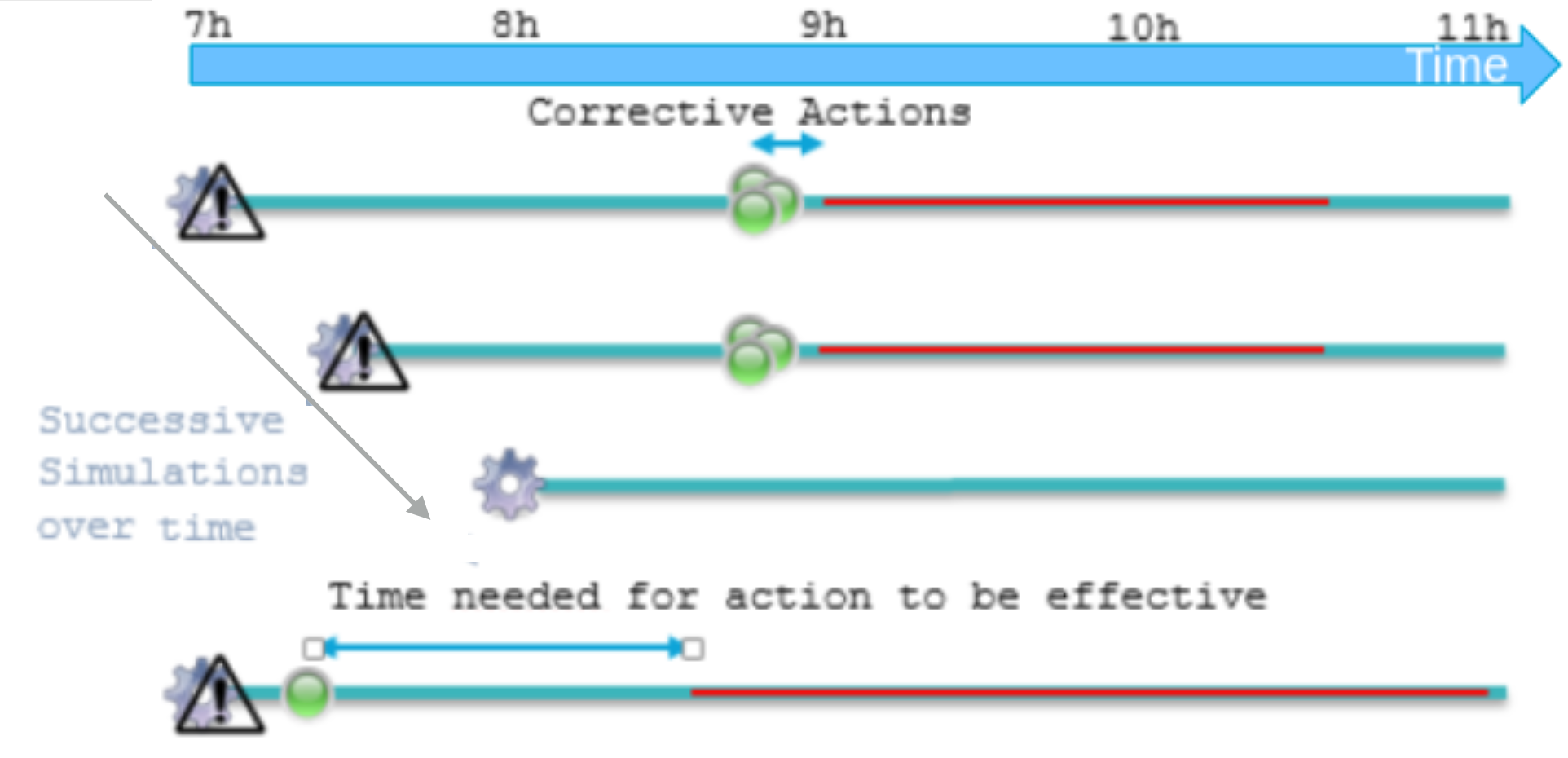}
\caption{A simple scenario where incidents are forecasted in multiple parts of a grid, and several corrective actions with different setup duration are possible. Green dots each represent an available action, and the blue arrows the time it needs to be effective. For instance, starting a nuclear power plant takes longer than applying topological grid changes, and would have a longer blue arrow.}
\label{fig:L2RPNserieDiff}
\end{figure}
A couple of questions arise on the best strategy to follow:
\begin{itemize}
    
    \item Which decisions have priority? It seems that a preventive action on area 2 should be urgently taken, but maybe the forecast isn't that reliable yet. 
    
    \item When should we implement actions? Waiting for the last simulation at 8:00am in area 1 shows that the forecasted problem disappeared which is the best option.
    \item How does applying a preventive action on area 2 reflect on area 1? Would it lead to a less secure grid state? What coordination is required? Maybe there's a new outage operation in this area that isn't yet taken into account by the simulation.
    \item Which of the three corrective actions in area 1 should be taken? The operator has to mediate between economical, practical and safety arguments, each with a degree of uncertainty over an activation horizon. 

    
    
    
\end{itemize}

Our objective here is not to provide any viable solution, but rather to demonstrate that grid operators are confronted with complex decisions even on apparently simple cases, in which context-dependent trade-offs always have to be made. Future works could build a library of such canonical cases to be studied in the community.

\subsection{Unified Interface \& Data collection as an industrial stack}
The hypervision framework relies on a generic single interface that should be able to integrate any kind of tasks, and apply to different industrial systems for instance. 
An example of an existing framework is the open-source Operator Fabric \cite{op_fab}. It could be used both by industrial and researchers as a unified interface for decision-making processes across teams. 
Such a framework is also a cornerstone to digitalize the decision-making process, centralize every necessary information, hence capitalize on them. This historical data collection is essential for continuous improvement, experiments, as well as for creating datasets from which AI can learn recommendations. Data should get labeled and its quality properly monitored. These developments should create a necessary technical stack or data platform for an assistant, to overcome challenges in deploying AI \cite{paleyes2020challenges,baier2019challenges}.

\subsection{Power system AI modules for assistant functions}\label{sec:AIModules}
Recent surveys list interesting developments of AI for climate change \cite{rolnick2022tackling} and for power systems more specifically \cite{kezunovic2020big}, \cite{duchesne2020recent}. For an assistant, AI can today be used to make corrective action recommendations to an operator through adaptive interpretable expert system \cite{marot2018expert},  imitation learning \cite{donnot2017introducing} or reinforcement learning on robustness, adaptability or trust dimensions \cite{marot2021learning,l2rpn_with_trust} which are key in Human-centered AI \cite{shneiderman2020human}. It can learn from user behaviour and help convey the consequences of the operator's action by comparison. Exhaustive risk assessment \cite{donnot2018optimization} also helps in prioritizing tasks. Further, automatic hierarchical and contextual representations of the grid \cite{marot2018guided} enable scope services and give greater flexibility to convey the right context and interpret a situation. \cite{crochepierre2020interpretable} also lets an AI learn interpretable and physically-consistent contextual indicators associated with a particular operator's task or help build knowledge graphs \cite{huang2020knowledge}. Finally, \cite{marot2019interpreting,boudjeloud2016interactive,gkorou2020get} let operators explore interactively and iteratively historical explainable factors across similar situations and decisions for augmenting and keeping up-to-date the system knowledge and proper labels. This should all be carefully developed within a trustworthy framework \cite{stiasny2022closing}. This is an illustrative sample of today's AI potential \cite{Li18, Shi20} to provide effective assistance functions and interactions which needs to be developed further. 

\subsection{Assistant evaluation \& development of shared benchmarks}
 In order to assess the relevance of an assistant in a real-world scenario, and eventually compare multiple assistants, it is necessary to set up repeatable evaluation protocols and define common benchmark tasks.
 
 As for now, there is not yet a standard testing protocol to evaluate AI assistants. However, we could draw insights from other domains such as interpretable machine-learning \cite{doshi2017towards}, explainable AI \cite{pruthi2020evaluating, rosenfeld2021better}, or interactive visualization \cite{Borgo2018InformationVE}. As done in \cite{doshi2017towards}, we could come to structured step-by-step experimental practices to evaluate candidate assistants on incrementally difficult tasks.
 Moreover, several Virtual-Assistant (VA) related studies have also tried to define custom evaluation criteria. In \cite{martin2020daphne}, authors compare their VA against both a simpler interactive data-exploration scheme and a non-interactive solution-search. They assess the use of their assistant on three factors: \textit{performance} with task-specific metrics; \textit{human-learning} using questions and tests at the end of each task; \textit{usability}  for example using System Usability Scale \cite{Brooke1996SUSA}. During this evaluation biases, over-reliance and deskilling (from sections \ref{sec:human_biaises} and \ref{sec:pitfalls}) can be addressed.

Because of confidentiality issues, it is often hard to share real-world data on decision-making problems. Thus, we should aim at developing synthetic but realistic environments from which to extract representative and relevant scenarios
, for example by drawing inspiration from road safety decision-making assessment frameworks \cite{dilemmas}.
Moreover, synthetic and realistic environments for sequential decision-making have recently been developed for power grids \cite{marot2020learning,marot2021learning,l2rpn_with_trust} and could be interactively further studied with grid2viz \cite{grid2viz} and grid2grame \cite{grid2game} study tools. 


\section{Conclusion}
In this paper, we have presented the framework and principles for designing an AI assistant based on the concept of Hypervision and bidirectional interactions for power grid operators. It combined insights from various research fields, opening new research directions for augmented decision-making. We have provided initial guidelines of expected functions and practical needs, as well as already available materials in power grids, to continue exploring this promising and very much needed new field of human-machine partnership. Building on this proposed framework and existing L2RPN with Trust environment \cite{l2rpn_with_trust}, future work should aim soon at extending testbed environments and instantiate first complete implementations of assistant to be benchmarked, in order to operate the grid with greater flexibility and coordination to support the ongoing energy transition.

\bibliographystyle{ieeetr}
\bibliography{powertech}

\end{document}